%% file: 00-main.tex
\pdfoutput=1
\documentclass[letterpaper, 10 pt, conference]{ieeeconf}  % Comment this line out
\IEEEoverridecommandlockouts                              % This command is only
\usepackage{cite}
\usepackage{amsmath,amssymb,amsfonts}
\usepackage{algorithmic}
\usepackage{graphicx}
\usepackage{textcomp}
\usepackage{lscape}
\usepackage{graphicx}
\usepackage{lipsum}% http://ctan.org/pkg/lipsum
\usepackage{graphicx}% http://ctan.org/pkg/graphicx
\title{\LARGE \bf
Deep-Learning-based Automated Palm Tree Counting and Geolocation in Large Farms from Aerial Geotagged Images
}

\title{\LARGE \bf Deep-Learning-based Automated Palm Tree Counting and Geolocation in Large Farms from Aerial Geotagged Images }

\author{Adel Ammar$^{1}$, Anis Koubaa$^{{1},{2}}$ 
\thanks{$^{1}$Prince Sultan University, Saudi Arabia. }
\thanks{$^{2}$CISTER, INESC-TEC, ISEP, Polytechnic Institute of Porto,Portugal.}
\thanks{This  work  is  supported by the Robotics and Internet-of-Things Lab at Prince Sultan University.}
}

\begin{document}

% \makeatletter
% % \IEEEoverridecommandlockouts
% %%%%%%%%%%%%%%%%%%%%%%%%%%%%% User specified LaTeX commands.
% \def\ps@IEEEtitlepagestyle{%
%   \def\@oddfoot{\mycopyrightnotice}%
%   \def\@evenfoot{}%
% }
% \def\mycopyrightnotice{%
%   {\footnotesize978-1-7281-2747-7/20/\$31 \textcopyright2020 IEEE\hfill} % <--- Change here
%   \gdef\mycopyrightnotice{} % just in case
% }

\maketitle

\thispagestyle{empty}
\pagestyle{empty}
% \IEEEpubid{\makebox[\columnwidth]{978-1-7281-2747-7/20/\$31 \textcopyright{}2020 IEEE \hfill} \hspace{\columnsep}\makebox[\columnwidth]{ }}

%%%%%%%%%%%%%%%%%%%%%%%%%%%%%%%%%%%%%%%%%%%%%%%%%%%%%%%%%%%%%%%%%%%%%%%%%%%%%%%%
\begin{abstract}
 
In this paper, we propose a deep learning framework for the automated counting and geolocation of palm trees from aerial images using convolutional neural networks. For this purpose, we collected aerial images in a palm tree Farm in the Kharj region, in Riyadh Saudi Arabia, using DJI drones, and we built a dataset of around 10,000 instances of palms trees. Then, we developed a convolutional neural network model using the state-of-the-art, Faster R-CNN algorithm.
Furthermore, using the geotagged metadata of aerial images, we used photogrammetry concepts and distance corrections to detect the geographical location of detected palms trees automatically.  This geolocation technique was tested on two different types of drones  (DJI  Mavic  Pro,  and  Phantom  4  pro), and was assessed to provide an average geolocation accuracy of 2.8m. This GPS tagging allows us to uniquely identify palm trees and count their number from a series of drone images, while correctly dealing with the issue of image overlapping. Moreover, it can be generalized to the geolocation of any other objects in UAV images\footnote{This is version 1 of the paper}.

\end{abstract}


\input{01-introduction.tex}

\input{02-related-works.tex}

\input{04-experimental.tex}

%\input{05-conclusion.tex}

\section*{ACKNOWLEDGMENT}
This work is supported by the Robotics and Internet-of-Things Lab of Prince Sultan University.

\bibliographystyle{ieeetr}
\bibliography{biblio,aniskoubaa_publication,biblio_new}
% \bibliography{aniskoubaa_publication,biblio_new}

\end{document}

%% file: 01-introduction.tex
\section{INTRODUCTION}

Tree counting from aerial images is a challenging problem with many applications such as forest inventory, crop estimation, and farm management. Nevertheless, counting the number of palms trees in large farms has been a challenging problem for agriculture authorities due to a massive number of trees and the inefficiency of manual counting approaches. The problem becomes even more laborious and tedious when we also need to identify the GPS location of palm trees for governance purposes. The inefficiency of traditional methods leads to inconsistent data collection about the number of palms, as reported by agriculture experts. 
For this purpose, we present a deep learning framework for building an inventory of individual palm trees by automatically counting and geolocating them using areal color images collected by unmanned aerial vehicles.
The remaining of the paper is organized as follows. Section II discusses the related works that dealt with car detection and aerial image analysis using CNN, and some comparative studies applied to other object detectiors. Then, section III describes the datasets and the obtained results. 

% \begin{figure}
% \begin{center}  
% \includegraphics[width=8cm]{images/introduction-figures/hajj_vms.jpg}
% \caption{\small \sl Surveillance Dashboard in Makkah during Pilgrimage: 5000 cameras and 100000 policemen.
% \label{fig:RPN_training}}  
% \end{center}  
% \end{figure} 

% \subsection{Motivation}

% \textit{Palm oil is the largest vegetable oil in the world in terms of produced volume, and 75\% of global production is used for food and cooking purposes. Sustainable management of the producing areas calls for the frequent assessment of field conditions.}

% \textit{...build an inventory of individual oil-palm trees using areal color images collected by unmanned aerial vehicles}

%% file: 02-related-works.tex
\section{RELATED WORKS}

% MAKE A SUMMARY TABLE

There have been several research studies on aerial image processing using deep learning in general, and on palm detection and counting more specifically. Before the area of deep learning, in 2011, Shafri et al. proposed in \cite{Helmi2011IJRS} a detection technique of the oil-palm tree by combining several techniques, namely, edge enhancement, spectral and blob analysis, and segmentation. Reference \cite{Li2016RemoteSensing} represents the first work to detect and count palm trees from multispectral QuickBird satellite images using deep learning, dated from 2016. The spatial resolution is 2.4m, but images were processed with the panchromatic band to achieve a 60 cm resolution. The authors developed a convolutional neural network detection with a sliding window approach to localize and classify palm trees in Malaysia with an accuracy of 96\%. It was shown that the proposed CNN detector provides better performance as compared to local maximum filter and template matching. In \cite{Li2017IGARSS}, the same authors proposed a classification technique based on AlexNet for the detection of palm trees from high-resolution satellite images. The classification accuracy achieved was 92\% to 97\% for the study area of palm tree farms of Malaysia. Our work differs from \cite{Li2017IGARSS} in several aspects. First, we consider high-resolution aerial images rather than satellite images, which provides a higher-resolution of 2 cm/pixel and more apparent features of palm trees. Besides, aerial images provide more up-to-date data as compared to their satellite counterparts. Second, we addressed an object detection problem rather than a classification problem, which requires both the localization and the classification of palm tree instances in images. Third, we do not only detect palm trees, but we also determine its geolocation from geotagged images.

In \cite{Zortea2018IGARSS}, the objective was to devise a deep learning algorithm for automatically building an inventory of palm trees from aerial images collected by drones. Their contribution was to combine the output of two CNN algorithms where the first is applied to 10 cm/pixel images to learn fine-grain features, and the second neural network 20 cm/pixel to focus on more coarse grain features. The authors achieved detection accuracy values between  91.2 and 98.8\% using the orthomosaic of decimeter spatial resolution. Our work differs in several aspects. First, we consider higher-resolution UAV-based aerial images of 2 cm/pixel. We also develop palm detection models based on state-of-the-art algorithms, namely, YoloV3, Faster RCNN, and EfficienDet. Furthermore, we use the metadata of geotagged images to identify the geolocation of detected palms. In our work, we can achieve an accurate inventory of palm tree farms not only in terms of counting but also for the trees' geolocalization. 

Some other works considered other types of trees such as \cite{Neupane2019PLOS} that developed a deep learning model for detecting banana trees from aerial images. They have reached the accuracy values of 96.4\%, 85.1\%, and 75.8\% for the altitudes 40, 50, and 60 meters, on the same farm. They have applied deep learning detection algorithm on orthomosaic maps. They have used Faster RCNN with a 42-layered Inception-v2 model feature extractor. Our work improves over \cite{Neupane2019PLOS} in that it can be applied to geotagged images, which enables us to uniquely identify each palm tree by its geolocation, and correctly deal with the issue of overlapping images.

%% file: 04-experimental.tex
 \section{Experiments}
 
  \begin{figure}[ht]
\centering
\includegraphics[width=8 cm]{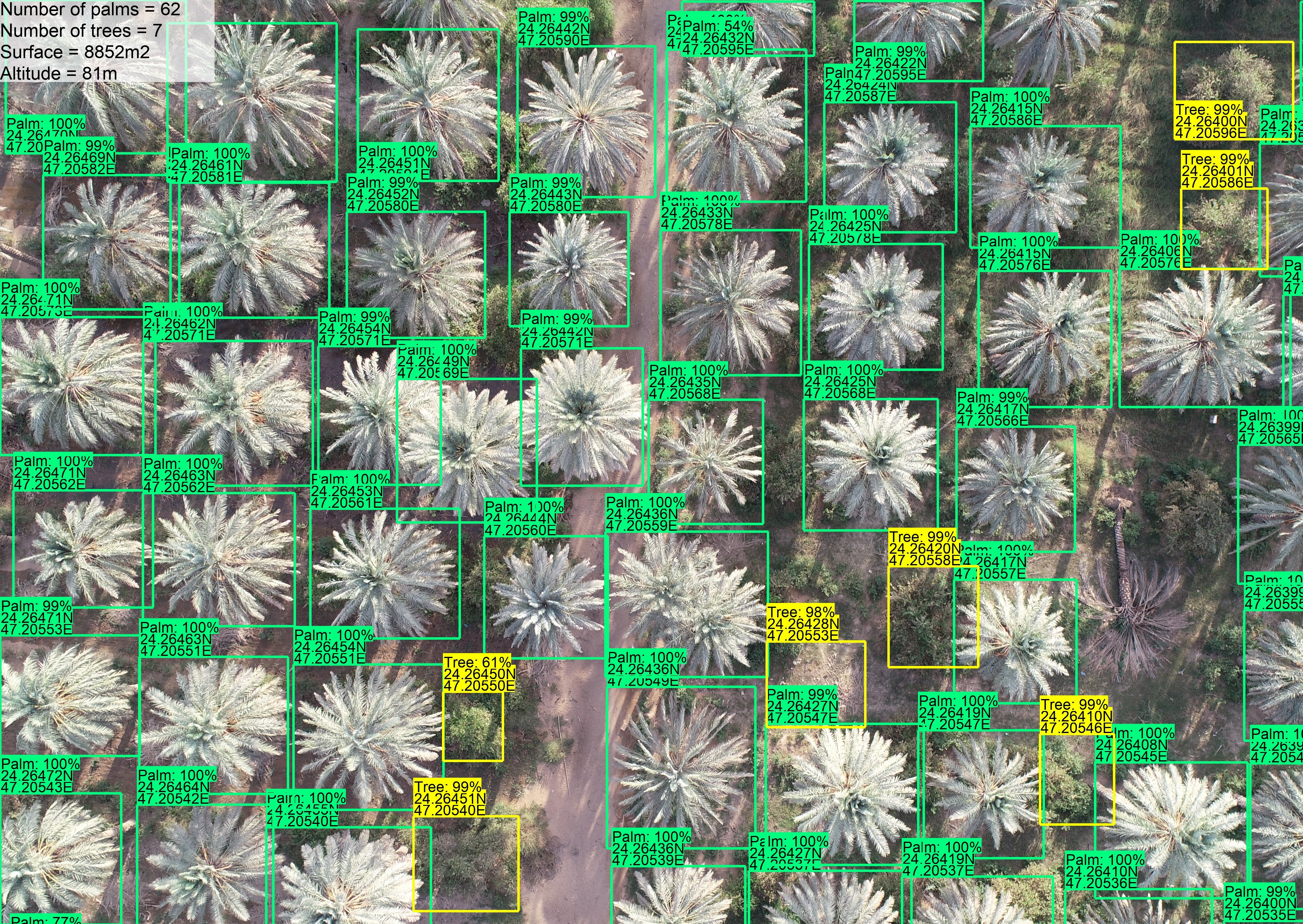} 
\caption{\small \sl Sample image of the palm counting and geolocation application.}
\label{fig:sample_detection_geolocation}
\end{figure}

 \begin{figure}[ht]
\begin{center}  
\includegraphics[width=7cm,height=6cm]{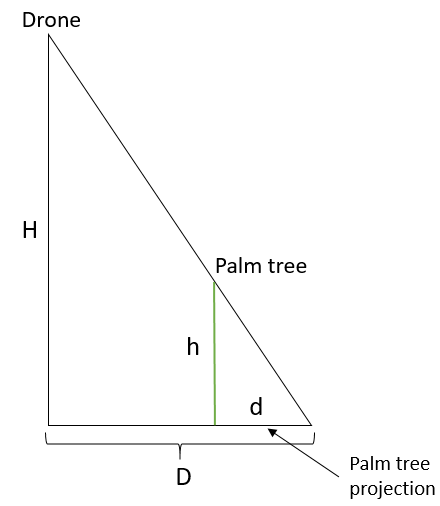}
\end{center}
\includegraphics[width=8.5cm]{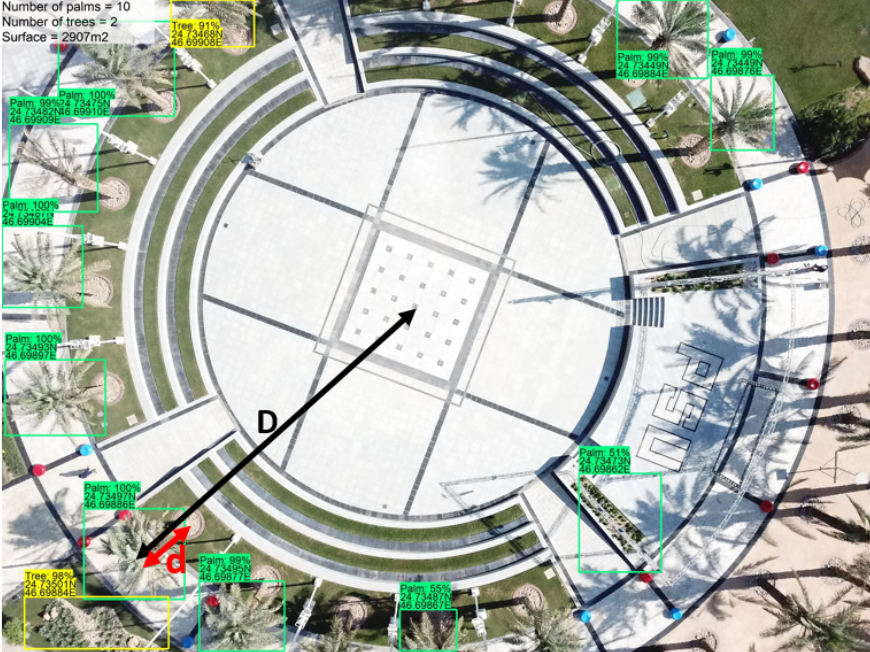}
\setlength{\belowcaptionskip}{-0.2cm}
\caption{\small \sl Distance correction (d) for the geolocation of palm trees from drone images. H is the drone altitude, h is the (average) palm tree height, D is the distance between the image center (drone's vertical projection on earth) and the position of the palm tree summit in the image (supposedly corresponding to the detected bounding box center).
\label{fig:distance_correction}}  
 
\end{figure}

 While the detection of palm infestation is a significant determinant for accurate crop estimation, it needs to be accompanied by a reliable tree counting technique, given the fact that traditional counting methods are inaccurate and inefficient. In order to address this problem, we propose an automated tree counting technique from aerial images, which can also have many other applications such as forest inventory, and farm management.  For this aim, we collected a dataset of 217 UAV images taken in a palm tree farm in Kharj region, in Saudi Arabia, with a total of 9,873 instances (8,652 palms, and 1,221 other trees). We manually labeled this dataset using Labelbox \cite{Labelbox}. Then, we trained a Faster R-CNN model \cite{Faster_R-CNN_journal}, which is a state-of-the-art two-stage object detection algorithm, on 80\% of the dataset (174 images). We obtained a precision of 94\% and a recall of 84\% for the detection of palm trees, on the testing dataset (43 images). The average precision (AP) at an IoU (Intersection over Union) threshold of 0.5 attains 83\%.

 \par Furthermore, we developed an algorithm that tags each detected tree with its GPS location by applying photogrammetry concepts to the metadata extracted from drone images (altitude and GPS location of the drone, image size, calibrated focal length, yaw degree), then applying a distance correction based on the ratio between the drone altitude and the estimated average palm height.  This geolocation technique was tested on two different types of drones (DJI Mavic Pro, and Phantom 4 pro), and was assessed to provide an average geolocation accuracy of 2.8m, a maximum of 4.9m, and a standard deviation of 1.2m. Figure \ref{fig:sample_detection_geolocation} shows an example of palm detection and geolocation in a UAV image, displaying on top of each detected bounding box the class of each object (palm tree or other tree), the classification confidence level, and the latitude and longitude of the bounding box center. 
 
The GPS tagging allows to uniquely identify, track and count the number of palm trees from a series of drone images, while correctly dealing with the issue of image overlapping while the drone is flying. This procedure can be generalized to the gelocation of any other objects in UAV images.
 
 To geolocate a pixel ($x,y$) in a drone image, We first calculate the equivalent distance to the central pixel ($x_c, y_c$) in the image frame:
 \[d_x=\frac{(x-x_c)}{F_c}H\]
 \[d_y=\frac{(y_c-y)}{F_c}H\]
 
 Where $H$ is the drone altitude, and $F_c$ is the calibrated focal length of the camera.
 
 Then, we apply a rotation by the value of the flight yaw, to convert $(d_x,d_y)$ to local tangent plane (LTP) coordinates. Finally, we apply a distance correction (Figure \ref{fig:distance_correction}) based on the average palm tree height to take account of projection issues that induce a difference between the position of the palm tree summit on the image (which corresponds to the center of the detected bounding box) and its footprint.

%  \subsection{Datasets}

% \begin{table}[]
% \caption{Number of images and instances in the training and testing datasets.}
% \label{tab:Dataset}
% \begin{tabular}{lcc}
% \hline
%                                  & \multicolumn{1}{l}{Training dataset} & \multicolumn{1}{l}{Testing dataset} \\ \hline
% Number of images                 &                    174               & 43                                  \\ \hline
% Percentage                       & ...\%                               & \%                              \\ \hline
% Instances of class "Palm tree"    &                                   &                                   \\ \hline
% Instances of class "Other tree"      &                                   &                                   \\ \hline
% \end{tabular}
% \end{table}

%  \subsection{Object detection}
%  \subsection{Geolocation accuracy}
%  - Comparison with Google maps and GPS locations measured by the drone on land.
 
%  - Accuracy of Google maps GPS locations (ref).